%% file: tdp2023.tex
\documentclass[runningheads,a4paper]{llncs}

\usepackage{ifxetex}
\ifxetex%
  \usepackage{fontspec}
  \usepackage{polyglossia}
  \setmainlanguage{english}
\else
  \usepackage[utf8]{inputenc}
  \usepackage[T1]{fontenc}
  \usepackage[english]{babel}
\fi

\usepackage{amssymb}
\usepackage{graphicx}
\usepackage{url}
\usepackage{float}
\usepackage{amsmath}
\usepackage{graphicx}
\usepackage{wrapfig}
\usepackage{booktabs}
\usepackage{multirow}
\usepackage{lipsum}
\usepackage{csquotes}
\usepackage{fancyhdr}
\usepackage{lastpage}
\usepackage[all]{nowidow}
\usepackage[inline]{enumitem}
\usepackage[usenames,dvipsnames]{xcolor}

\makeatletter
\def\Hline{%
\noalign{\ifnum0=`}\fi\hrule \@height 1pt \futurelet
\reserved@a\@xhline}
\makeatother

\title{Hibikino-Musashi@Home\\2023 Team Description Paper}

\author{
Tomoya Shiba
\and Akinobu Mizutani
\and Yuga Yano
\and Tomohiro Ono
\and Shoshi Tokuno
\and Daiju Kanaoka
\and Yukiya Fukuda
\and Hayato Amano
\and Mayu Koresawa
\and Yoshifumi Sakai
\and Ryogo Takemoto
\and Katsunori Tamai
\and Kazuo Nakahara
\and Hiroyuki Hayashi
\and Satsuki Fujimatsu
\and Yusuke Mizoguchi
\and Moeno Anraku
\and Mayo Suzuka
\and Lu Shen
\and Kohei Maeda
\and Fumiya Matsuzaki
\and Ikuya Matsumoto
\and Kazuya Murai
\and Kosei Isomoto
\and Kim Minje
\and Yuichiro Tanaka
\and Takashi Morie
\and Hakaru Tamukoh
}
\authorrunning{Tomoya Shiba et al.}
\institute{
Kyushu Institute of Technology\\
The University of Kitakyushu\\
\email{hma@brain.kyutech.ac.jp} \\
\url{https://www.brain.kyutech.ac.jp/~hma/}
}

\begin{document}
\maketitle

%
%

\begin{abstract}
This paper describes an overview of the techniques of Hibikino-Musashi@Home, which intends to participate in the domestic standard platform league. The team has developed a dataset generator for the training of a robot vision system and an open-source development environment running on a human support robot simulator. The robot system comprises self-developed libraries including those for motion synthesis and open-source software works on the robot operating system. The team aims to realize a home service robot that assists humans in a home, and continuously attend the competition to evaluate the developed system. The brain-inspired artificial intelligence system is also proposed for service robots which are expected to work in a real home environment.
\end{abstract}


\section{Introduction}
Hibikino-Musashi@Home (HMA) is a robot development team organized by students at the Kyushu Institute of Technology and the University of Kitakyushu in Japan. The team was founded in 2010 and has participated in the RoboCup@Home JapanOpen in the Open Platform League (OPL), Domestic Standard Platform League (DSPL), and Simulation-DSPL. It has been participating in the RoboCup @Home league since 2017 and intends to participate in RoboCup 2023 to present the outcome of the latest development and research. In addition to the RoboCup, the team has participated in the World Robot Challenge (WRC) 2018 and 2020 and the service robotics category of the partner robot challenge (real space). 
HMA focuses on the development of a robot vision system, particularly a dataset generation system for the training of an object recognition system. It also develops libraries for various tasks including object recognition, grasping point estimation, motion synthesis, and navigation, and releases them as open sources. The current focus of our attention is “environment understanding,” with which we aim to develop a robot that adapts to dynamic changes in an environment.


\section{Hardware Overview}

We use an external computer mounted on the back of the TOYOTA Human Support Robot (HSR) because the computational resources built into the HSR could not support the robot systems that include deep neural networks.
To extract the maximum performance of our system, we use a ThinkPad X1 Extreme Gen5 (Intel Core i9-12900H CPU, 32GB RAM, and NVIDIA GeForce RTX 3080Ti GPU 16GB). Consequently, the computer inside the HSR is used to run basic HSR software, such as sensor drivers, motion planners, and actuator drivers. Our system is complete within the mounted laptop and internal HSR computer; no external Internet connection is required to run the system. This increases the operational stability of the HSR.

\section{Software Overview}

\begin{figure}[bt]
\begin{center}
\includegraphics[scale=0.5]{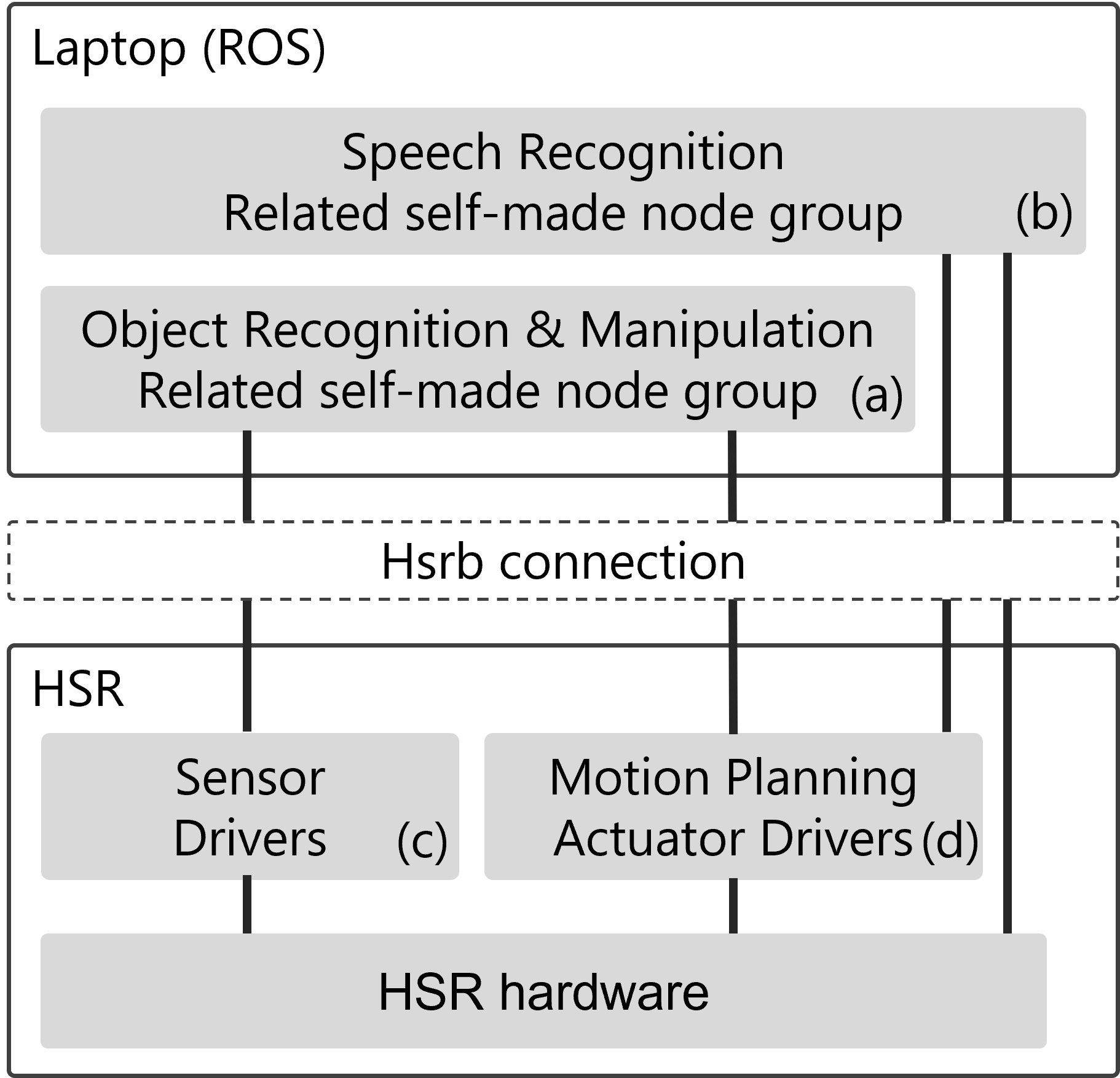}
\caption{Block diagram overview of our HSR system. [HSR, Human Support Robot; ROS, robot operating system]}
\label{fig:softOverview}
\end{center}
\end{figure}

This section explains the software system of the HSR. 
Figure \ref{fig:softOverview} presents an overview of the HMA's software systems for HSR \cite{yamamoto2019robomech}. The system comprises an object recognition and manipulation system (a), a speech recognition system (b) in a laptop computer mounted on the HSR, and the sensor and actuator drivers (c) and (d), respectively, on the internal HSR computer.


\subsection{Object Recognition}
The object recognition system is one of the most fundamental units of a robot system. The recognition result is utilized to determine the object class and grasping points. We use an instance segmentation-based system, YolactEdge (optimized method of You Only Look At CoefficienTs, YOLACT \cite{bolya2019ICCV} for edge devices) \cite{yolactedge-icra2021}.

To train the object recognition system based on a deep neural network, a large number of training datasets is required. Moreover, a complex annotation phase is necessary to create labels and instance masks of objects.
In the RoboCup@Home competition, predefined objects are typically announced during the setup period before the start of the competition. Thus, preparing the dataset rapidly to train the recognition system is necessary.

We develop a dataset generation system based on the PyBullet \cite{coumans2016pybullet} simulator to prepare the dataset to train the recognition system \cite{ono2022ar}. 
This system places 3D models of objects in a 3D environment and automatically creates labels and masks. 
We use EinScan-SP \cite{einscan}, a compact 3D scanner, to create 3D models of objects on a turntable.

\begin{figure}[tb]
\begin{center}
\includegraphics[width=0.75\columnwidth]{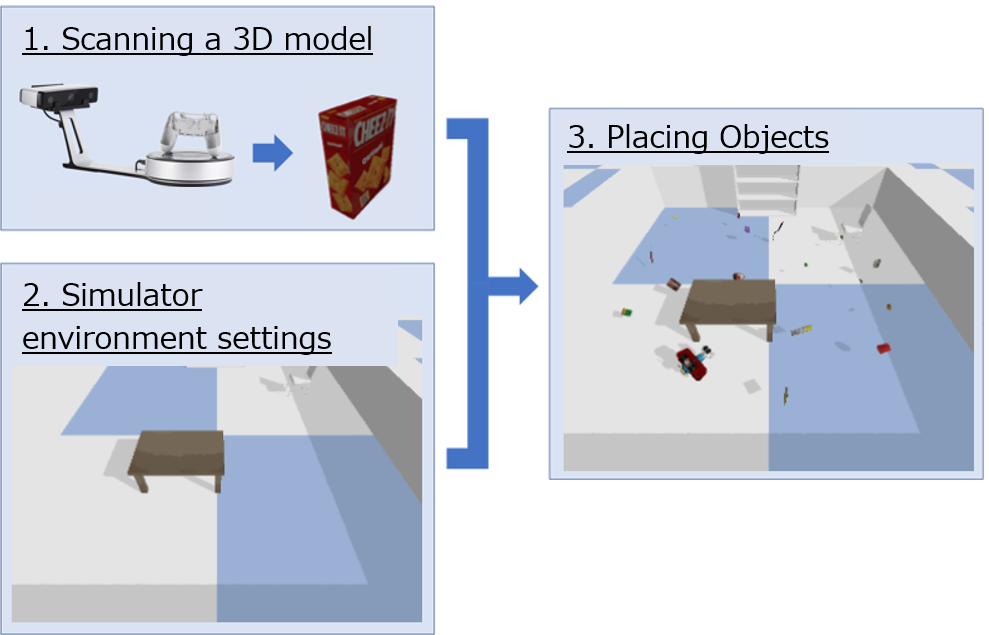}
\caption{Placing scanned objects on a 3D simulator.}
\label{fig:annotation2}
\end{center}
\end{figure}

As shown in Fig. \ref{fig:annotation2}, the scanned 3D objects are spawned on the 3D environment, and the objects and environment are shot from various angles to create the dataset. The light conditions, placement of furniture, and the texture of the background (floor, wall, and ceiling) are changed to randomize a domain at each shot.
The placement of objects is random, but the range of possible placement areas is defined for each task. Annotation data for the training images can be generated automatically because object labels and positions are known.
Generating 500,000 images requires less than 2 h (using a six-core CPU in parallel).

The recognition system can be trained only using the dataset generated by this system; no additional training is necessary to adapt to each real environment. Figure \ref{fig:annotation2} shows the processing flow to generate training images for YolactEdge.

\subsection{Motion Synthesis}
We improve the speed of the HSR from two viewpoints: motion synthesis and software processing speed.

We combine several motions to reduce waiting time. For instance, considering a situation where the robot picks up an object with its arm and moves to the storage place, in the conventional method, the robot has to wait until the end of the previous arm movements to move to the target point. The robot arm and base can be driven separately. We optimized the robot's motion by controlling each joint of the arm during navigation. By adopting this method, the robot can move to the next action without waiting for the completion of the previous motion. 

Regarding the processing speed, we aim to operate all software at 30 Hz or higher, which is the operation cycle of the camera and sensors.
To reduce the waiting time for software processing, which causes the robot to stop, the essential functions, such as object recognition and object grasping-point estimation, of home service robots need to be executed in real-time. We optimized the processing flow and algorithms of the robot functions.

\begin{figure}[bt]
  \begin{center}
    \includegraphics[scale=0.65]{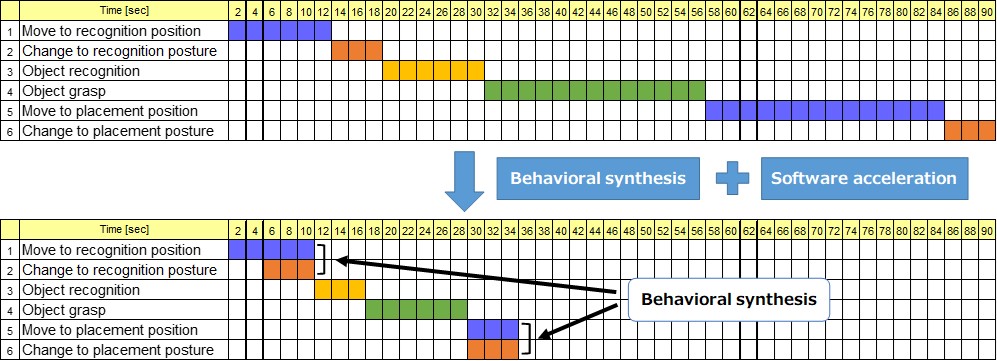}
    \caption{Speed comparison between a conventional system and the proposed motion synthesis system.}
    \label{fig:synthesis}
  \end{center}
\end{figure}

We used these two optimization methods for the tidy-up task in the WRC 2018 and 2020 (Fig. \ref{fig:synthesis}). 
In the WRC 2018 and 2020 results, for which we won first place, our achieved speedup was approximately 2.6 times the prior record. Our robot could tidy up within 34 s per object; therefore, we expect that it can tidy up approximately 20 objects in 15 min. 
We will utilize this technique for all tasks of picking up objects for more common household environments such as shelves and desks.

\subsection{Speech Recognition}
In a task involving human--robot interaction such as asking for a guest's name or order, the robot needs to recognize human speech.
In the competition, these names are announced before the task execution. We create a name list for human and object names to improve the accuracy of the speech recognition system. Because the Internet connection of the competition site is occasionally unstable, we use an offline speech recognition system, VOSK\cite{vosk}, to maintain the stability of the robot system.
In a competition site, the noise contained in a guest's speech should be suppressed. A strong nose reduction system is required to suppress the noise from spectators or the voices of other robots.
We apply the noise reduction system proposed by Sainburg et al. \cite{tim2020noisereduce} before the voice recognition process. 

Occasionally, the speech recognition system fails to recognize human speech. In such a scenario, the robot uses a QR code shown by humans to interact with humans instead of speaking. Moreover, we developed a QR code generator that works offline to assist human--robot interaction\cite{qr_generator}.

\subsection{Environment Understanding}
\begin{figure}[tb]
  \begin{center}
    \includegraphics[width=0.5\columnwidth]{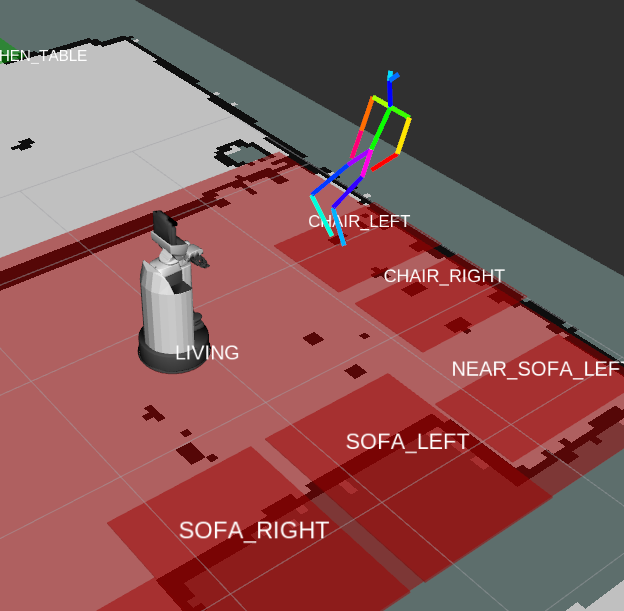}
    \caption{Example of a semantic map}
    \label{fig:semantic-map}
  \end{center}
\end{figure}

A semantic map is required for the autonomous task execution of service robots. As shown in Fig. \ref{fig:semantic-map}, we add semantic information using a JavaScript Object Notation (JSON) file to a pre-acquired environment map created using Real-Time Appearance-Based Mapping (RTAB-map)\cite{mathieu2019rtabmap}.
Before the task execution, we define areas of room and furniture using a JSON file and utilize that information to determine the semantic location of the robot or humans in an arena.

Based on the semantic and obstacle information obtained by sensors, the robot dynamically generates an action using navigation systems. For example, in the task where the human requests the robot to move to the room specified by name, the robot utilizes semantic information. In a real home environment, the robot can generate efficient action using the memory acquired through experiences rather than generating action every time. We proposed a navigation system based on the robot's experiences and action generation system using a brain-inspired artificial intelligence model\cite{tanaka2019biai,mizutani2021,nakagawa2022}.

To estimate human attributes, such as gender or dress, we use class-specific residual attention (CSRA)\cite{Zhu2021CSRA}. Using the estimated attributes, the robot can identify humans in a room. By combining the semantic information and names dictated by humans, the robot can identify humans in a room by estimated attributes. This enables the introduction of guests' information to family members or offering food or drinks according to requests. 
In a real environment, robots are expected to serve foods and drinks according to preferences, which should be acquired or estimated by  human--robot interactions. We proposed family preference acquisition and estimation systems using a brain-inspired amygdala model\cite{tanaka2020amygdala}.

\section{Re-usability}
Open-source HSR simulators\cite{tmc_wrs_docker,hsrb_robocup_dspl_docker} are the official platforms of the virtual competition used in RoboCup 2021.
Our system developed for this competition is available on GitHub (\url{https://github.com/Hibikino-Musashi-Home/hma_wrs_sim_ws}) with documentation and sample programs. The system includes the motion synthesis technology and object recognition described in the previous section. Notably, the tidy-up task in WRC can be developed even without the physical HSR. We believe this contributes to increasing the population who research and develop home service robots.

The developed system is divided into several ROS packages. The packages can be applied to other robots by changing the topic. Thus, the system is not limited to the HSR, that is, it can be used for a wide range of applications that works on ROS.


\section{Conclusions}
This paper described the techniques for creating an intelligence system of home service robots. The automatic dataset generation system is essential to train the visual system of the service robot in a limited time. We integrated self-developed and open-source systems for task execution. Moreover, our team proposed a brain-inspired artificial intelligence system for service robots that are expected to work in a real home environment. The robot's function with the proposed brain-inspired AI was demonstrated in an open challenge and finals.
In future work, we will automate the process to create an environment map that can facilitate adapting to dynamic changes in the environment.

\section*{Acknowledgment}
This paper is based on results obtained from a project, JPNP16007, commissioned by the New Energy
and Industrial Technology Development Organization (NEDO).
This paper is also  supported by Joint Graduate School Intelligent Car, Robotics \& AI,
Kyushu Institute of Technology student project, YASKAWA electric corporation project,
JSPS KAKENHI grant number 22K17968.

\bibliography{ref}
\bibliographystyle{unsrt}

\newpage
\input{RobotDescription}

\end{document}

%% file: RobotDescription.tex
\section*{Appendix 1: Robot's Software Description}
\begin{wrapfigure}[20]{r}{0.25\textwidth}
        \centering
        \includegraphics[width=0.25\textwidth]{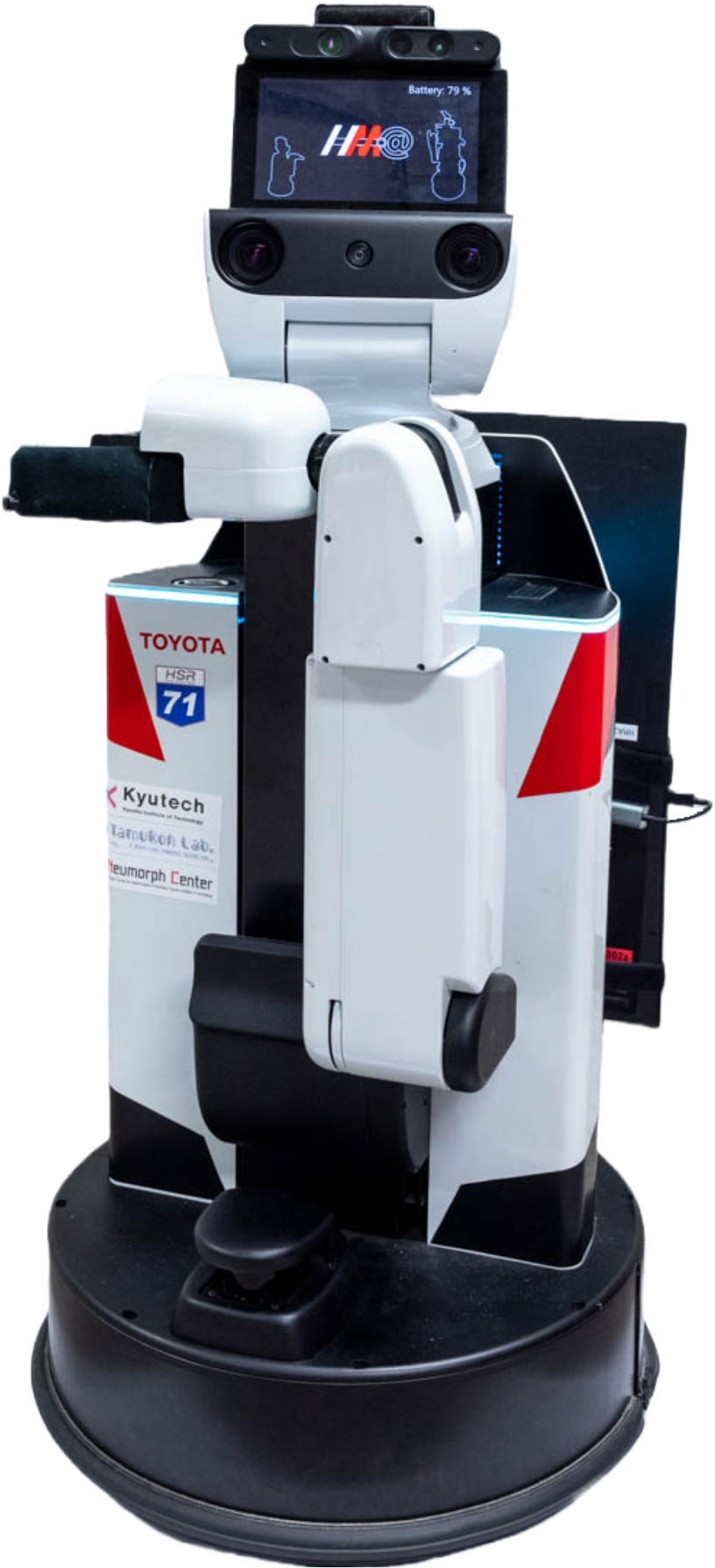}
        \caption{HSR}%
        \label{fig:hsr}
\end{wrapfigure}

The following is the software stack of our robot system, shown in Fig. \ref{fig:hsr}.
\begin{itemize}
	\item OS: Ubuntu 20.04.
	\item Middleware: ROS Noetic.
	\item State management: SMACH (ROS).
	\item Speech recognition: vosk \cite{vosk}.
	\item Object detection: YolactEdge \cite{yolactedge-icra2021}.
	\item Human detection / action recognition:
		\begin{itemize}
			\item Depth image + particle filter.
			\item Lightweight OpenPose \cite{Daniil2018lwop}.
			\item tracking person Deep Sort \cite{Wojke2018deep}
		  \item action recognition ST-GCN \cite{stgcn2018aaai}
		\end{itemize}
	\item Attribute recognition: CSRA \cite{Zhu2021CSRA}.
	\item SLAM: rtabmap \cite{mathieu2019rtabmap} (ROS).
	\item Path planning: move\_base (ROS).
\end{itemize}

The following is the specification of the laptop computer mounter on our HSR.
\begin{itemize}
        \item Model name: ThinkPad X1 Extreme Gen5
        \item CPU: Intel Core i9-12900H
        \item RAM: 32GB
        \item GPU: NVIDIA GeForce RTX 3080Ti (16GB)
\end{itemize}

\section*{Appendix 2: Competition results}

\begin{table}[t]
\begin{center}
\caption{Results of recent competitions. [DSPL, domestic standard-platform league; JSAI, Japanese Society for Artificial Intelligence; METI, Ministry of Economy, Trade and Industry (Japan); OPL, open-platform league; RSJ, Robotics Society of Japan]}
\label{tab:result}
\begin{tabular}{l|l|l} \hline
	\multicolumn{1}{c|}{Country} & \multicolumn{1}{c|}{Competition} & \multicolumn{1}{c}{Result} \\ \hline \hline

	Japan & RoboCup 2017 Nagoya & {\bf @Home DSPL 1st} \\
        				      && @Home OPL 5th \\ \hline

        Japan & RoboCup Japan Open 2018 Ogaki & @Home DSPL 2nd \\
        				                      && \textbf{@Home OPL 1st} \\
                                        	         && JSAI Award \\ \hline

        Canada & RoboCup 2018 Montreal & \textbf{@Home DSPL 1st} \\
                                         && P\&G Dishwasher Challenge Award \\ \hline

        Japan & World Robot Challenge 2018 & {\bf Service Robotics Category} \\
                                              && \textbf{Partner Robot Challenge Real Space 1st} \\
                                         	 && METI Minister's Award, RSJ Special Award \\ \hline

        Australia & RoboCup 2019 Sydney & @Home DSPL 3rd \\ \hline

        Japan & RoboCup Japan Open 2019 Nagaoka &\textbf{@Home DSPL 1st} \\
        				                          && \textbf{@Home OPL 1st} \\ \hline
        Japan & RoboCup Japan Open 2020 & @Home Simulation Technical Challenge 2nd \\
                                        && \textbf{@Home DSPL 1st} \\
                                        && @Home DSPL Technical Challenge 2nd \\
                                        && \textbf{@Home OPL 1st} \\
                                        && \textbf{@Home OPL Technical Challenge 1st} \\
                                        && @Home Simulation DSPL 2nd \\ \hline
        Worldwide & RoboCup Worldwide 2021 & @Home DSPL 2nd \\
                                        && \textbf{@Home Best Open Challenge Award 1st} \\
                                        && \textbf{@Home Best Test Performance: } \\
                                        && \textbf{    Go, Get It! 1st} \\
                                        && \textbf{@Home Best Go, Get It! 1st} \\ \hline
        Japan & World Robot Challenge 2020 & \bf{Service Robotics Category} \\
                                        && \bf{Partner Robot Challenge Real Space 1st} \\ \hline
        Japan & RoboCup Asia-Pacific 2021 Aichi Japan & \textbf{@Home DSPL 1st}\\
                                        && \textbf{@Home OPL 1st} \\ \hline
        Japan & RoboCup Japan Open 2021 & \textbf{@Home DSPL 1st} \\
                                        && \textbf{@Home DSPL Technical Challenge 1st} \\
                                        && @Home OPL 2nd \\
                                        && \textbf{@Home OPL Technical Challenge 1st} \\ \hline
        Thailand & RoboCup 2022 Bangkok & @Home DSPL 3rd\\ \hline
\end{tabular}
\end{center}
\end{table}

Table \ref{tab:result} shows the results achieved by our team in recent competitions.
We have participated in the RoboCup and World Robot Challenge for several years. Our team has won several prizes and academic awards. \par

\section*{Appendix 3: Links}

\begin{itemize}
  \item Team Video \\ \url{https://youtu.be/E2S1UxdXR0k} \\
  \item Team Website \\ \url{https://www.brain.kyutech.ac.jp/~hma} \\
  \item GitHub \\ \url{https://github.com/Hibikino-Musashi-Home} \\
  \item Facebook \\ \url{https://www.facebook.com/HibikinoMusashiAthome} \\
  \item YouTube \\ \url{https://www.youtube.com/@hma_wakamatsu}
\end{itemize}